\RequirePackage{amsmath}
\documentclass[runningheads]{llncs}
\usepackage[T1]{fontenc}
\usepackage{graphicx}
\usepackage{color}
\usepackage{etoolbox} 

\usepackage{amsmath,amssymb,amsfonts}
\usepackage{graphicx}
\usepackage{amsmath}
\usepackage{amssymb}
\usepackage{enumerate}
\usepackage{multirow}
\usepackage{threeparttable}

\usepackage[pagebackref=false,breaklinks=true,colorlinks,bookmarks=false]{hyperref}
\usepackage{times}
\usepackage{array}
\usepackage{verbatim}
\usepackage{algorithm}
\usepackage{algpseudocode}
\usepackage{lettrine}
\usepackage{booktabs}
\begin{document}
\bibliographystyle{plain}

\title{DASSF: Dynamic-Attention Scale-Sequence Fusion for Aerial Object Detection
}
%
%
\author{Haodong Li \inst{1} \and
Haicheng Qu\inst{1*}}
\institute{School of Software, Liaoning Technical University, Huludao, China}

\maketitle              
\begin{abstract}
The detection of small objects in aerial images is a fundamental task in the field of computer vision. Moving objects in aerial photography have problems such as different shapes and sizes, dense overlap, occlusion by the background, and object blur, however, the original YOLO algorithm has low overall detection accuracy due to its weak ability to perceive targets of different scales. In order to improve the detection accuracy of densely overlapping small targets and fuzzy targets, this paper proposes a dynamic-attention scale-sequence fusion algorithm (DASSF) for small target detection in aerial images. First, we propose a dynamic scale sequence feature fusion (DSSFF) module that improves the upsampling mechanism and reduces computational load. Secondly, a x-small object detection head is specially added to enhance the detection capability of small targets. Finally, in order to improve the expressive ability of targets of different types and sizes, we use the dynamic head (DyHead). 
 The model we proposed solves the problem of small target detection in aerial images and can be applied to multiple different versions of the YOLO algorithm, which is universal. Experimental results show that when the DASSF method is applied to YOLOv8, compared to YOLOv8n, on the VisDrone-2019 and DIOR
datasets, the model shows an increase of 9.2\% and 3.0\% in the mean average precision (mAP), respectively, and outperforms the current mainstream methods.

\end{abstract}
\section{Introduction}
\vspace{-0.3cm}

Object detection is now widely applied across various fields, such as smart transportation~\cite{jiao2023dsam}, medical diagnosis~\cite{ni2018multiple}, industrial manufacturing~\cite{zhao2018surface}, and re-identification~\cite{shen2023git}. 
With the ongoing advancements in UAV technology and the maturation of remote sensing technology, aerial image target detection has emerged as a significant focus due to its vast potential. However, compared to traditional target detection scenarios, aerial images present unique challenges, including a wide range of target scales, small object sizes, diverse angle variations, complex backgrounds, and susceptibility to solar radiation and atmospheric transparency. These factors significantly complicate the accurate detection and identification of small targets in aerial images. Therefore, further research is needed to develop innovative methods and technologies to enhance the accuracy and efficiency of small target detection in these contexts.

According to the MS COCO dataset~\cite{lin2014microsoft} definition in the field of object detection, small objects are those smaller than 32 × 32 pixels. CFIL~\cite{weng2023cross} introduces a frequency-domain feature extraction module~\cite{weng2024enhancing} and frequency domain feature interaction to enhance salient features. MFC~\cite{qiao2022novel} proposes a frequency-domain filtering module to achieve dense target feature enhancement. LR-FPN~\cite{li2024lr} improves remote sensing target detection by enhancing low-level positional information and fine-grained context interaction, leading to superior performance in agricultural and urban planning applications. PBSL~\cite{shen2023pbsl} introduces a multimodal alignment approach to highlight relevant features and suppress irrelevant information. To address the issues of blurred boundaries and significant scale variations in high-resolution aerial images, Wang et al.~\cite{wang2023improved} made enhancements to the aerial image target detection task based on the YOLOX-X algorithm. By incorporating data augmentation, introducing a path aggregation network (PANet), and adding x-small detection heads, they significantly improved the overall detection capability of the network.

While certain results have been achieved in aerial scene target detection tasks using these methods, there are still some problems, such as insufficient extraction of features of different scales by the neck network, information loss during the transmission process of small target features, and the use of too many cascade networks or attention mechanisms resulting in high computational overhead. In order to solve these problems, we present a YOLO model with dynamic-attention scale-sequence fusion (DASSF) to significantly improve detection accuracy. The main contributions of this paper can be summarized as follows:

\begin{itemize}
    \item [\textcolor{black}{$\bullet$}]We design a dynamic scale sequence feature fusion (DSSFF) module to accurately and efficiently extract global high-level semantic information in images of different scales, which not only reduces the model complexity but also can accurately detect small objects through point sampling.
    \item[\textcolor{black}{$\bullet$}]By adding a x-small detection head with a resolution of 160 × 160, we can more
exactly identify the categories of small targets and more precisely locate the coor-
dinates of small targets during the detection stage, thereby improving the detection
accuracy of models for small targets.
    \item [\textcolor{black}{$\bullet$}]We use a dynamic head (DyHead) that combines scale, spatial, and task attention to improve the ability to integrate full-scale object features during the detection phase. It improves the detection of occluded targets commonly found in drone images and the blurring of targets affected by weather in remote sensing images.
    \item [\textcolor{black}{$\bullet$}]The DASSF method we finally propose not only significantly enhances the detection accuracy of objects at different scales and categories in aerial images, but also can be flexibly applied to various YOLO series algorithms.

\end{itemize}

\section{Proposed Method}\label{sec:method} 
\vspace{-0.3cm}
\subsection{The Overall Architecture of DASSF-YOLO}
\begin{figure}[t]
\centering
\includegraphics[width=12cm,height=6cm]{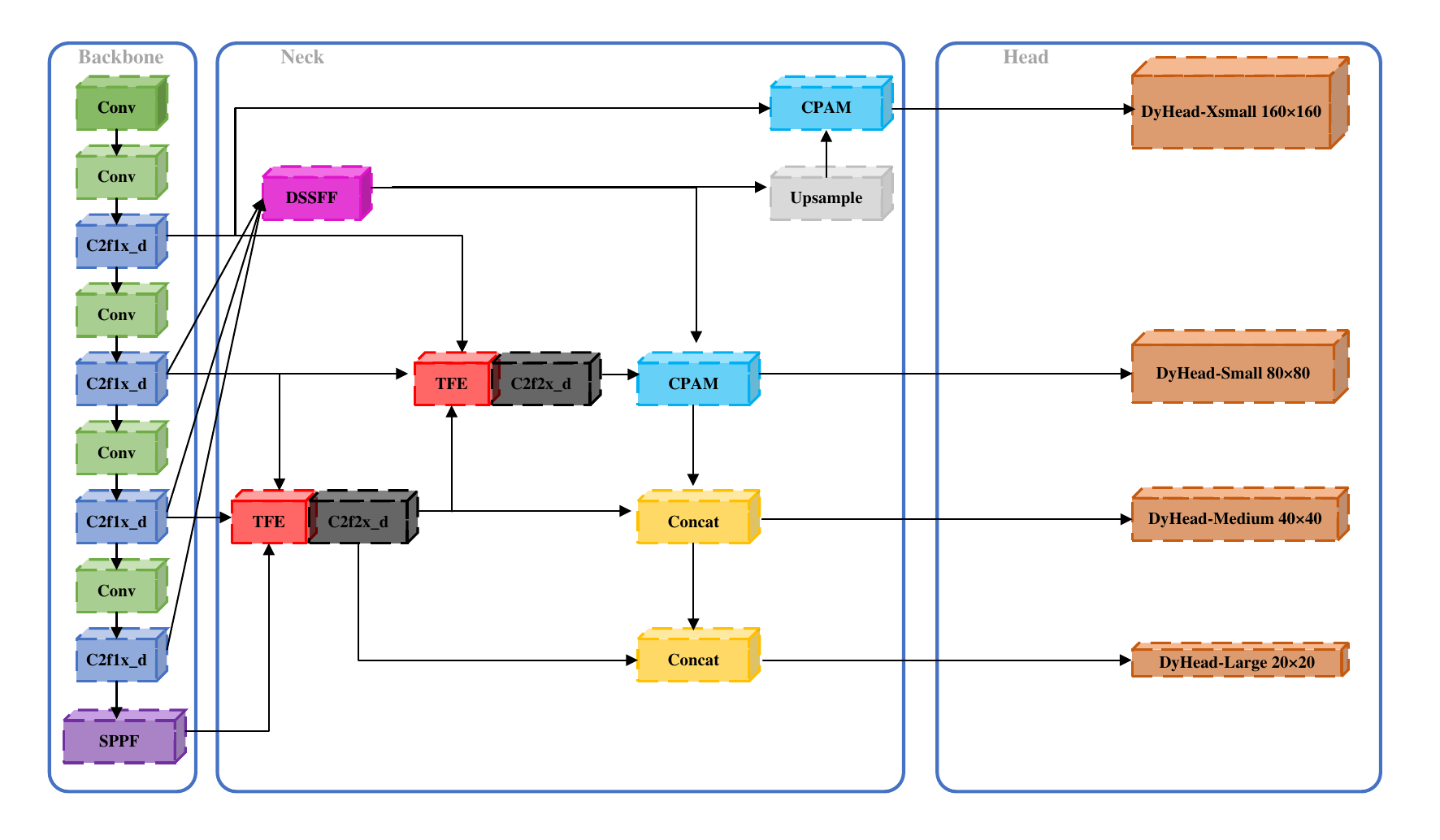}
\caption{The overall architecture of DASSF. DSSFF is dynamic scale sequence feature fusion module, TFE is triple feature encoding module, CPAM is channel and position attention mechanism, and DyHead is dynamic detection head.}
\label{fig:DASF-YOLOv8-new}
\end{figure}

\vspace{-0.3cm}
Fig \ref{fig:DASF-YOLOv8-new} shows the overall architecture of the dynamic-attention scale-sequence fusion algorithm 
(DASSF) network model we designed. We use CSPDarknet53 as the backbone network. We utilize ASF-YOLO~\cite{kang2023asf}'s neck network to enhance the detection of small, dense, and blurry targets in aerial images. The neck of the network utilizes the TFE in ASF-YOLO twice for fusing feature maps of different dimensions to obtain rich global channel information. Then apply the DSSFF to obtain accurate local location information. Then, the processed information is effectively interacted through the CPAM. Finally, by adding the x-small detection head and applying the DyHead. The model can detect objects of various scales in aerial images.

\subsection{Improvements to the Neck}  
\vspace{-0.3cm}
\subsubsection{Triple Feature Encoding Module}
The TFE module is a feature fusion mechanism. First, adjust the number of channels of the large, medium and small feature layers to make them equal through CBS operation. Then, the large-scale feature map is subjected to a down-sampling operation of maximum pooling + average pooling, which helps to retain the high-resolution features and the diversity of semantic information of different objects in aerial images; for small-scale feature maps, the nearest neighbor interpolation method is used for upsampling, which can maintain the richness of local features of low-resolution images and prevent the loss of small target location feature information. Finally, feature maps of different scales are fused through concat operations.

 \vspace{-0.3cm}

\subsubsection{Dynamic Scale Sequence Feature Fusion Module}

The original SSFF module was designed for the P3 layer and is a key component used to process multi-scale information and has the ability to extract features of different scales.
Scale means detail in the image. A blurry image may lose details, but the structural features of the image can be preserved, helping to solve the image blur problem in satellite remote sensing images. The input image of SSFF is in formula \ref{equ:SSFF1}.

\begin{equation}
\scriptsize
    \text{F}_{\text{o}}(\text{w}, \text{h})=\text{G}_{\text{o}}(\text{w}, \text{h}) \times \text{f}(\text{w}, \text{h}) .
    \label{equ:SSFF1}
\end{equation}

\begin{figure}[t]
\centering
\includegraphics[width=12cm]{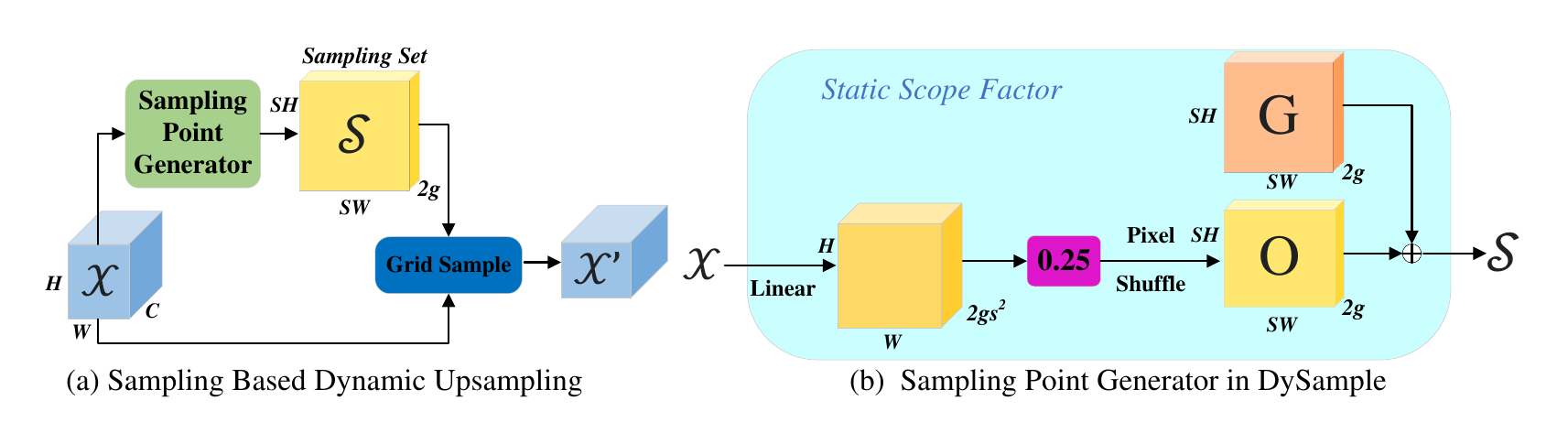}
\caption{The structure of the DySample. S represents the upsampling ratio, G represents the grid sampling point coordinates, and O represents the point position offset generated by the dynamic sampling point generator. SH and SW represent respectively sampling height and width. $\text{g}\text{s}^2$ represents the number of channels of the feature map after passing through the linear layer.}
\label{fig:DySample}
\end{figure}

Where f(w, h) represents a 2D input image with width w and height h. $\text{G}_{\text{o}}\text{(w, h)}$ is the filter used for smooth convolution. This module contains two upsampling operations for the P4 and P5 layers. The nearest upsampling method in the original module will lead to the loss of key image details and requires a lot of calculation and parameter overhead. Therefore, we introduce DySample~\cite{liu2023learning}, an ultra-lightweight and effective dynamic upsampler to replace the nearest upsampling method in the original module. The structure is shown in fig \ref{fig:DySample}.

\begin{figure}[t]
\centering
\includegraphics[width=12cm,height=4cm]{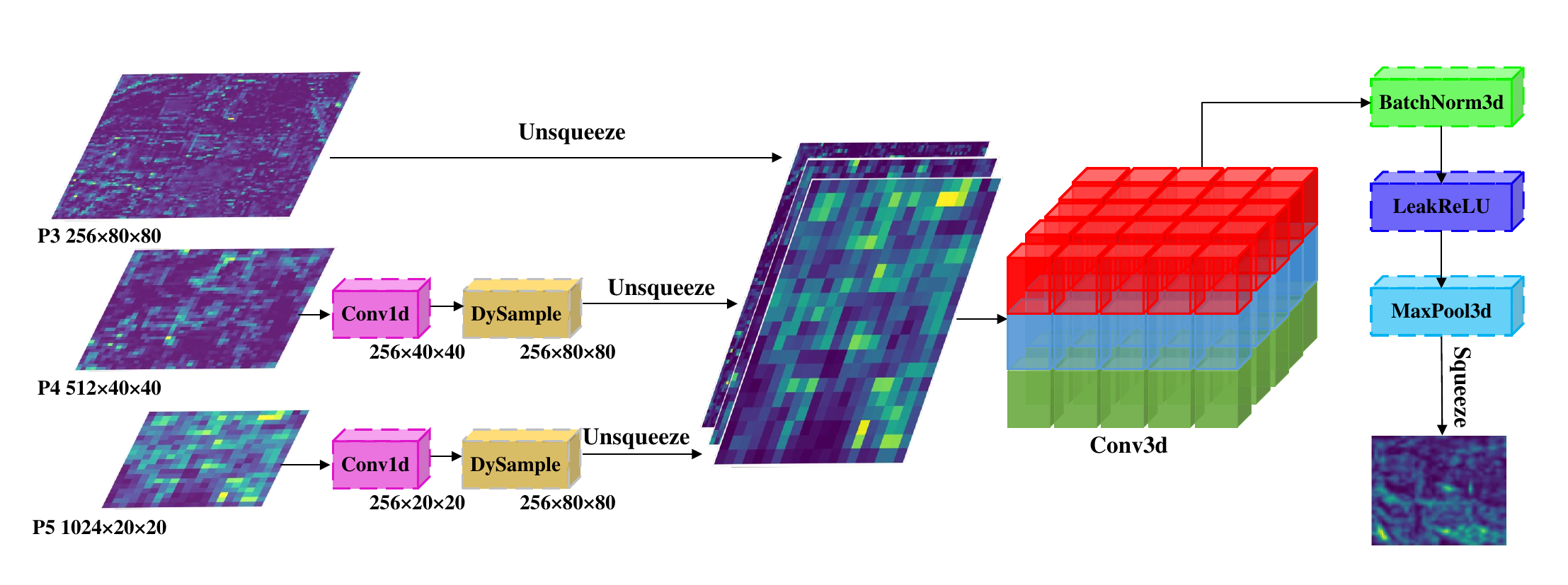}
\caption{The structure of the DSSFF. The features are extracted efficiently and accurately through dynamic upsampling, feature map stacking, and 3D convolution normalization activation operations. The detailed process of dynamic upsampling is shown in algorithm \ref{alg:1}.}
\label{fig:DSSFF}
\end{figure}

We adopt a static factor sampling method, based on the theory of adjusting point sampling position offset, to dynamically create a sampling set of point positions through the sample point generator in the feature map. Given the dimension C × H × W and the upsampling factor g, through the dynamic sampling generator, an upsampled feature map of size C × SH × SW is output. The structure of the DSSFF module is shown in fig \ref{fig:DSSFF}.
The pseudocode for dynamic upsampling with a static sampling factor of 0.25 is shown in algorithm \ref{alg:1}.

\begin{algorithm}[t]
\small
    \caption{The process of dynamic upsampling}
    \begin{flushleft}
    \hspace*{\algorithmicindent} \textbf{Input:} The feature map x of size C × H × W, upsampling multiple scale and groups default to 4. \\
    \end{flushleft}
    \begin{flushleft}
    \hspace*{\algorithmicindent} \textbf{Output:} The feature map x' of size C × scale × H × scale × W.\\
    \end{flushleft}
    \begin{algorithmic}[1]
    \State Generate a 2D convolutional layer for offset and perform normal distribution initialization, and get $\text{out}_1$.
    \State Generate the initial position for offset calculation, and get $\text{out}_2$.
    \State Apply $\text{out}_1$ to the input x and adjust the range by × 0.25, then append the offset of $\text{out}_2$, and then transform it into 2 × -1 × H × W, and get $\text{out}_3$.
    \State Create a normalized target coordinate grid, and get $\text{out}_4$.
    \State Add $\text{out}_3$ and $\text{out}_4$ and normalize them to the [-1,1] interval, and get $\text{out}_5$.
    \State Use pixel\_shuffle to upsample coordinates for $\text{out}_5$ and adjust the output size, then use grid\_sample for bilinear interpolation sampling, and get out.
    \State \Return out.
    \end{algorithmic}
    \label{alg:1}
\end{algorithm}

 \vspace{-0.3cm}

\subsubsection{Channel and Position Attention Mechanism}
 The CPAM integrates the DSSFF and TFE modules, which focus on information-rich channels and small object features related to spatial location. This allows the model to more accurately identify and locate small targets in images, thereby improving the detection capabilities of detailed small objects. Input 1 is the detailed features after TFE processing as channel attention network, which is used to effectively capture cross-channel interactions. This is an attention mechanism that does not require dimensionality reduction. The capture of local cross-channel interactions is achieved using 1D convolutions of size k, where the kernel size k represents the coverage of local cross-channel interactions. Using the output of the channel attention mechanism and the feature map processed by DSSFF as the input of the position attention network, the position information of different targets can be extracted.

\subsection{Improvements to the Head}  
\vspace{-0.3cm}
\subsubsection{X-small Head}
 Aerial image targets have a wide range of scales. Aerial images captured by drones include small objects such as pedestrians, bicycles, and motorcycles, and are characterized by dense targets and background occlusion. Remote sensing images taken by satellites include large objects such as ships, sites, bridges, etc., and the targets are relatively scattered. Therefore, in order to take into account the effective detection of objects of different scales, we added a 160 × 160 x-small target detection head to the YOLO algorithm. The final detection head includes P2, P3, P4, and P5.
 \vspace{-0.3cm}

\subsubsection{Dynamic Head}
The aerial imagery targets used in this study presented complex backgrounds. Because the target in the drone image is blocked by houses and trees and the target in the remote sensing image is affected by light and clouds. The scale of the target is easy to change, and the image is easy to become blurred and distorted. Therefore, it is crucial that the detection algorithm has a full range of perception capabilities. DyHead~\cite{dai2021dynamic} was proposed by Dai et al, which simultaneously combines scale-aware attention (\text{$\pi_\text{L}$}) in formula~\ref{equ:dyheadlf}, spatial-aware attention (\text{$\pi_\text{S}$}) in formula~\ref{equ:dyheadsf}, and task-aware attention (\text{$\pi_\text{C}$}) in formula~\ref{equ:dyheadcf}, enhances the model's adaptability to various target sizes, understanding of object placement, and context understanding. 
\begin{equation}
\scriptsize
    \pi_{\text{L}}(\text{F}) \cdot \text{F} = \sigma\left( f\left( \frac{1}{\text{S} \cdot \text{C}} \sum_{\text{S}, \text{C}} \text{F} \right) \right) \cdot \text{F} .
    \label{equ:dyheadlf}
\end{equation}
Scale-aware attention (\text{$\pi_\text{L}$}) performs average pooling on the input feature map, then uses a 1 × 1 convolution layer and ReLU activation function for feature extraction, then uses the hard-sigmoid function to balance model accuracy and speed, finally the elements are multiplied with the input feature map.
\begin{equation}
\scriptsize
    \text{\(\pi_{\text{S}}(\text{F}) \cdot \text{F}\)} = \frac{1}{\text{L}} \sum_{\text{l}=1}^{\text{L}} \sum_{\text{j}=1}^{\text{K}} \text{w}_{\text{l, j}} \cdot \text{F}\left( \text{l} ; \text{p}_{\text{j}}+\text{\(\Delta\)} \text{p}_{\text{j}} ; \text{c} \right) \cdot \text{\(\Delta\)} \text{m}_{\text{j}} .
    \label{equ:dyheadsf}
\end{equation}
Spatial-aware attention (\text{$\pi_\text{S}$}) first processes the input tensor using a 3 × 3 convolutional layer to obtain the offset value of the feature map and the weight term of the feature map offset, and then weights and sums all features.
\begin{equation}
\scriptsize
    \text{\(\pi_{\text{C}}(\text{F}) \cdot \text{F}\)}  = \text{\(\max\)} \left( \text{\(\alpha\)}^{\text{1}}(\text{F}) \cdot \text{F}_{\text{C}} + \text{\(\beta\)}^{\text{1}}(\text{F}), \text{\(\alpha\)}^{\text{2}}(\text{F}) \cdot \text{F}_{\text{C}} + \text{\(\beta\)}^{\text{2}}(\text{F}) \right) .
    \label{equ:dyheadcf}
\end{equation}
Task-aware attention (\text{$\pi_\text{C}$}) is first average pooling in the L × S dimension to reduce the number of channels. Subsequently, two fully connected layers are adopted and activated using the ReLU function and then passed through a normalization layer. Finally, different channel values are output according to different tasks to complete the task perception of the feature map.

\section{Experiment and Analysis}\label{sec:exp}  
\vspace{-0.3cm}
In order to validate the superiority of the proposed DASSF method, we combine it with YOLOv8n and conduct comparison and ablation experiments on two datasets. The comparison results show that our proposed method significantly improves the accuracy of small targets in aerial images.

\subsection{Datasets}
\begin{figure}[t]
\centering
\includegraphics[width=12cm]{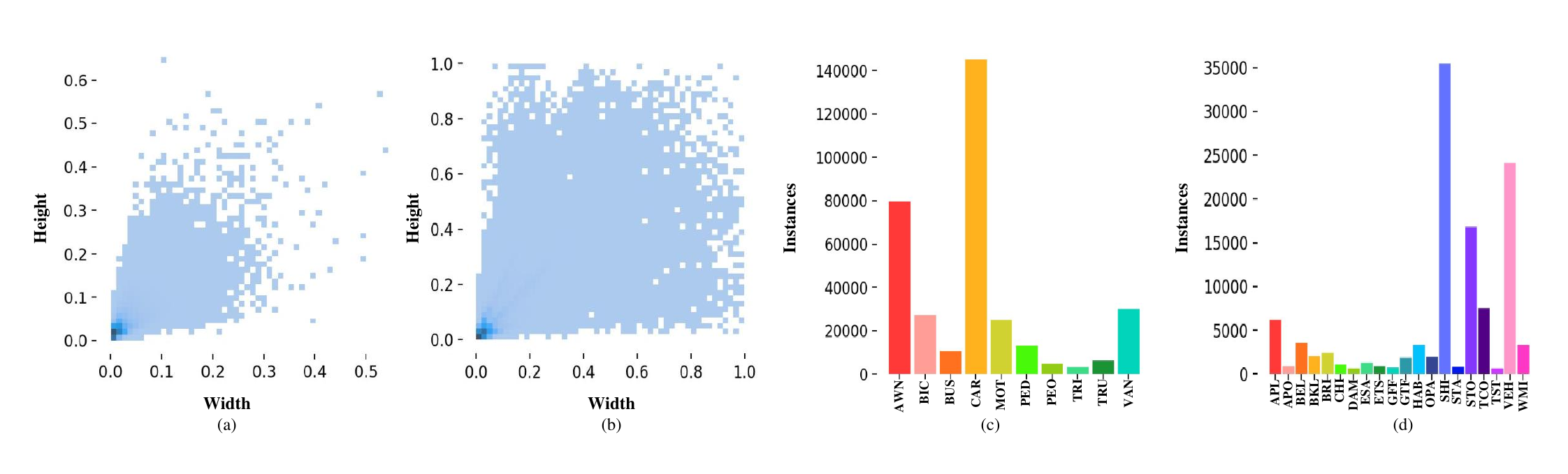}
 
\caption{(a) The size distribution of objects in VisDrone-2019 dataset; (b) The size distribution of objects in DIOR dataset; (c) The category distribution of objects in VisDrone-2019 dataset; (d) The category distribution of objects in DIOR dataset.}
\label{fig:label+hw}
\end{figure}

\vspace{-0.3cm}
In our experiments, we use two datasets. The first is the Tianjin University AISKYEYE team publicly released the VisDrone-2019~\cite{du2019visdrone} dataset. This dataset is designed for target detection in UAV images of remote sensing scenes with high diversity.  The images are annotated with labels for ten categories, including awning tricycle, bicycle, bus, car, motorcycle, pedestrian, person, tricycle, truck and van. The dataset is divided into three distinct subsets: 6471 images for training, 548 images for validation, and 1610 images for testing. The second is the DIOR~\cite{li2020object} remote sensing dataset was released by Northwestern Polytechnical University in 2018. This benchmark dataset contains 23,463 images and 192,472 instances for object detection in optical remote sensing images. It covers 20 common object categories, including different places and objects such as aircraft (APL), chimneys (CH), overpasses (OP), etc. The dataset is divided into three different subsets: 14077 images for training, 4694 images for validation, and 4692 images for testing. The size and category distributions of the two datasets are shown in fig \ref{fig:label+hw}.

\subsection{Implementation Details and Evaluation Metrics} 
\vspace{-0.3cm}
The hardware configuration of this experiment includes: CPU: AMD EPYC 7551P 32-Core Processor, GPU: NVIDIA RTX A4000, Memory: 16G. The software environment includes: Ubuntu 20.04.1, python 3.8.10, Torch 1.13.1. We train for 100 epochs with batchsize set to 8 and use stochastic gradient descent (SGD) to update parameters. In order to evaluate the model's performance, this experiment uses precision (P), recall (R), mean average precision (mAP), and frames per second (FPS) as indicators.

\subsection{Comparison with State-of-the-art Methods} 
\vspace{-0.3cm}
\subsubsection{Comparisons on VisDrone-2019}

We use the VisDrone-2019 dataset to conduct comparative experiments with mainstream two-stage and one-stage target detection algorithms, including SSD, Cascade R-CNN, RetinaNet, ATSS, YOLOv6n, DAMO-YOLO, RTMDET, YOLO-MS, Gold-YOLO.  The comparative experimental results in table \ref{visdrone_result} show that our proposed method surpasses the selected target detection method YOLOv-8n and improves the detection accuracy by 9.2\%. And it surpassed other mainstream target detection algorithms in the table, achieving new SOTA results of 39.6\% and 23.5\% in mAP50 and mAP50:95 respectively. The best detection accuracy was achieved in 8 out of 10 categories. Due to the addition of a small target detection head and the application of the DSSFF module, the model's ability to detect small targets and extract and fuse features of objects of different scales are enhanced. Excellent detection results can be achieved for large-sized targets such as buses and garbage trucks, as well as small-sized targets such as pedestrians and motorcycles.

\begin{table}[t]
\caption{Comparison of detection results of mainstream methods on the VisDrone-2019 dataset.}
\label{visdrone_result}
\centering
\setlength{\tabcolsep}{1.2pt}
\small
\begin{tabular}{|l| *{9}{c|}c|c|c|}
\hline
Method &  \textit{Awn} & \textit{Bic} & \textit{Bus} & \textit{Car} & \textit{Mot} & \textit{Ped} & \textit{Peo} & \textit{Tri} & \textit{Tru} & \textit{Van} & \scriptsize mAP50 (\%) &\scriptsize mAP50:95 (\%) \\
\hline
SSD  &  11.2 & 7.4 & 49.8 & 63.2 & 19.1 & 18.7 & 9.0 & 11.7 & 33.1 & 30.0 & 25.3 & 14.6\\
Cascade R-CNN    & 8.6 & 7.6 & 34.9 & 54.6 & 21.4 & 22.2 & 14.8 & 14.8 & 21.6 & 31.5 & 23.2 & 13.4\\
RetinaNet    & 4.2 & 1.4 & 17.8 & 45.5 & 11.8 & 13.0 & 7.9 & 6.3 & 11.5 & 19.9 & 13.9  & 8.1\\
ATSS     & 8.5 & \textbf{18.8} & 52.1 & 76.6 & 41.4 & 42.7 & 22.3 & 28.4 & \textbf{36.9} & 41.4 & 37.3  & 23.0\\
YOLOv6n    & 8.1 & 3.9 & 23.0 & 72.2 & 29.0 & 28.3 & 23.0 & 15.9 & 21.7 & 33.1 & 27.3  & 15.7\\
DAMO-YOLO    & 11.0 & 6.7 & 42.3 & 74.2 & 33.4 & 32.8 & 26.2 & 19.3 & 23.5 & 35.9 & 30.5  & 17.6\\
RTMDET   & 9.5 & 4.3 & 39.6 & 69.6 & 29.0 & 24.6 & 19.7 & 17.9 & 23.4 & 35.7 & 27.3  & 16.1\\
YOLO-MS    & 11.9 & 6.0 & 39.6 & 73.5 & 32.2 & 31.4 & 25.2  & 18.9 & 22.1 & 35.4 & 29.6 & 16.7\\
Gold-YOLO    &  11.5 & 6.9 & 44.3 & 74.3 & 33.7 & 32.3 & 26.3 & 20.1 & 26.6 & 36.9 & 31.3 & 18.0\\
\hline
Baseline   & 10.9 & 6.4 & 42.5 & 74.5 & 33.6 & 32.3 & 25.5 & 19.2 & 23.8 & 35.7 & 30.4 & 17.4\\
Ours   & \textbf{16.1} & 14.4 & \textbf{52.5} & \textbf{81.0} & \textbf{45.9} & \textbf{44.6} & \textbf{36.8} & \textbf{24.7} & 31.2 & \textbf{44.4} & \textbf{39.6} & \textbf{23.5}\\
\hline
\end{tabular}

\end{table}

\vspace{-0.3cm}

\subsubsection{Comparisons on DIOR}

Furthermore, we also compare the proposed method with mainstream methods on the DIOR dataset.  As can be seen from table \ref{dior_result}, our proposed method surpasses other mainstream target detection methods in terms of overall accuracy, reaching 84.4\%. Exceeding RetinaNet, ATSS, and RTMDET by 4.3\%, 1.3\%, and 9.1\% respectively on mAP50. Due to the use of DyHead with self-attention, the model's ability to perceive objects of different scales in remote sensing images is enhanced.
 Moreover, the DSSFF module solves the problem of misdetection and leakage of targets such as chimneys and windmills that are affected by light, clouds and other factors. And due to the improved upsampling mechanism in the scale sequence feature fusion module, which reduces the amount of calculation, the FPS exceeds the seven target detection methods in the table, ensuring the real-time performance of the proposed method.
\begin{table*}[t]{\textwidth=0mm}
\caption{Comparison of detection results of mainstream methods on the DIOR dataset.}
\label{dior_result}
\centering
\setlength{\tabcolsep}{3pt}
\small
\begin{tabular}{|l |l|c| c|  }

\hline
Method   & Year & mAP50 (\%) &mAP50:95 (\%)  \\
\hline

SSD ~\cite{liu2016ssd}  & 2016 & 72.6 & 41.9 \\
Cascade R-CNN ~\cite{cai2018cascade}  &2017 & 80.9 & 58.1 \\
RetinaNet ~\cite{lin2017focal}  &2018 & 80.1 &53.2 \\
ATSS  ~\cite{zhang2020bridging} &2020  & 83.1 & 52.1 \\
YOLOv6n ~\cite{li2023yolov6} &2022  & 79.1  & 56.8 \\
DAMO-YOLO ~\cite{xu2022damo} &2022  & 82.5  & 59.0 \\
RTMDET ~\cite{lyu2022rtmdet} &2022  & 75.3  & 55.9 \\
YOLO-MS ~\cite{chen2023yolo} &2023  & 81.8  & 57.8 \\
Gold-YOLO ~\cite{wang2024gold} &2023  & 82.7  & 59.1 \\
\hline
Baseline  &2023  & 81.4 & 57.2 \\
Ours   &-  & \textbf{84.4} & \textbf{60.6} \\
\hline
\end{tabular}

\end{table*}

\subsection{Ablation Studies and Analysis}  

\vspace{-0.3cm}
We conduct ablation experiments on the proposed method on two datasets, and the experimental results are shown in table \ref{ablation_result}. Baseline is YOLOv8n. The detection accuracy index of the baseline model is at the lowest position. The mAP50 and mAP50:95 of the finally proposed improved model on the two datasets increased by 9.2\%, 6.1\%, 3.0\% and 3.4\% respectively compared with the baseline model. This shows that the improved model has a slight increase in calculation volume and inference time due to the addition of attention and x-small target detection heads, but can significantly improve detection performance. For the improvement of DSSFF, mAP50 and mAP50:95 increased by 0.9\% and 0.4\% respectively in the VisDrone-2019 dataset. On the DIOR dataset, mAP50 increased by 0.8\%, but mAP50:95 decreased by 0.3\%. This shows that the feature fusion mechanism of DSSFF is more helpful for identifying dense small targets. For the improvement of x-small target detection head, mAP50 and mAP50:95 increased by 2.7\% and 1.7\% respectively in the VisDrone-2019 dataset. In the DIOR dataset, mAP50 increased by 0.2\%, and mAP50:95 decreased by 1.6\%. This shows that the x-small target detection head is more conducive to detecting aerial datasets that mainly contain small targets, but it is not effective in detecting aerial datasets with evenly distributed large and small scale targets. For the improvement of DyHead, the two datasets mAP50 and mAP50:95 increased by 1.9\%, 1.2\%, 1.0\% and 1.3\% respectively, while FPS also decreased. This demonstrates that DyHead, which combines size, space, and task attention simultaneously, enhances the model's overall expressiveness, but it also increases computational effort and inference time.  As for the pairwise combination of three improvements, as reflected in mAP50 and mAP50:95, further accuracy improvement can be achieved on the basis of a single improvement. On the DIOR dataset, when the improved DSSFF and x-small target detection head are added, the accuracy index is improved compared with the final proposed method, but this does not affect the trend of improving the overall detection accuracy.

\begin{table}[t]
\centering
\setlength{\tabcolsep}{1.3pt}
\small
\caption{Ablation study on VisDrone-2019 and DIOR dataset.}
\label{ablation_result}
\begin{tabular}{|c|c|c|c|c|c|c|c|c|}
\hline

\multicolumn{1}{|c|}{\scriptsize Dataset } & \multicolumn{4}{c|}{VisDrone-2019} & \multicolumn{4}{c|}{DIOR} \\
\hline
  \scriptsize Method & \scriptsize Precision  & \scriptsize Recall & \scriptsize mAP50 (\%) & \scriptsize mAP50:95 (\%) & \scriptsize Precision & \scriptsize Recall  &\scriptsize  mAP50 (\%)& \scriptsize mAP50:95 (\%)  \\
\hline
\scriptsize Baseline  & 40.8 & 31.0 & 30.4 & 17.4   
         & 86.5 & 74.8 & 81.4 & 57.2   \\
\scriptsize DSSFF   & 41.2 & 32.3 & 31.3 & 17.8  
   & 85.5 & 75.7 & 82.2 & 56.9   \\
\scriptsize X-small   & 44.1 & 32.9 & 33.1 & 19.1   
    & 85.0 & 75.5 & 81.6 & 55.6   \\
\scriptsize DyHead   & 42.3 & 32.7 & 32.3 & 18.6   
   & 85.6 & 75.8 & 82.4 & 58.5   \\
\scriptsize DSSFF+X-small   & 44.7 & 34.2 & 34.2 & 20.0   
     & \textbf{87.1} & 76.6 & 82.7 & 58.9   \\
\scriptsize DSSFF+DyHead   & 42.2 & 32.7 & 32.9 & 19.1   
     & 86.7 &  76.9 & 83.3 & 59.9   \\
\scriptsize X-small+DyHead    & 47.5 & 37.9 & 38.1 & 22.5   
      & 86.3 & 77.1 & 83.4 & 57.9   \\

\hline
\scriptsize Ours   & \textbf{49.9} & \textbf{39.0} & \textbf{39.6} & \textbf{23.5}   
     & 85.6 & \textbf{77.8} & \textbf{84.4} & \textbf{60.6}   \\
\hline
\end{tabular}
\end{table}

We conduct variant experiments on the proposed DSSFF module using different upsampling methods on the VisDrone-2019 dataset. As can be seen from the experimental results in table \ref{variant}, the upsampling method using DySample is optimal in all four accuracy indicators and has the highest FPS. Compared with the original nearest sampling method, precision, recall, mAP50 and mAP50:95 increase by 1.2\%, 1.0\%, 0.7\% and 0.5\% respectively. FPS reaches 46.7. This demonstrates that the DSSFF module we implement in the model not only enhances detection performance but also decreases computational overhead.
\begin{table}[t]
\centering
\setlength{\tabcolsep}{2.6pt}
\small
\caption{Experimental results of DSSFF module and different upsampling methods on VisDrone-2019 dataset.}
\label{variant}
\begin{tabular}{|l|c|c|c|c|c|}
\hline
Method & Precision & Recall & mAP50 (\%) & mAP50:95 (\%)  & FPS \\
\hline
Bilinear & 49.2 & 36.7 & 38.2 & 22.6  & 34.1 \\
CARAFE~\cite{wang2019carafe} & 49.3 & 37.5 & 38.9 & 22.9  & 35.3 \\
Nearest & 48.7 & 38.0 & 38.9 & 23.0  & 45.8 \\
DySample (Ours) & \textbf{49.9} & \textbf{39.0} & \textbf{39.6} & \textbf{23.5}  & \textbf{46.7} \\
\hline
\end{tabular}
\end{table}

\begin{figure}[t]
\centering
\includegraphics[width=12cm]{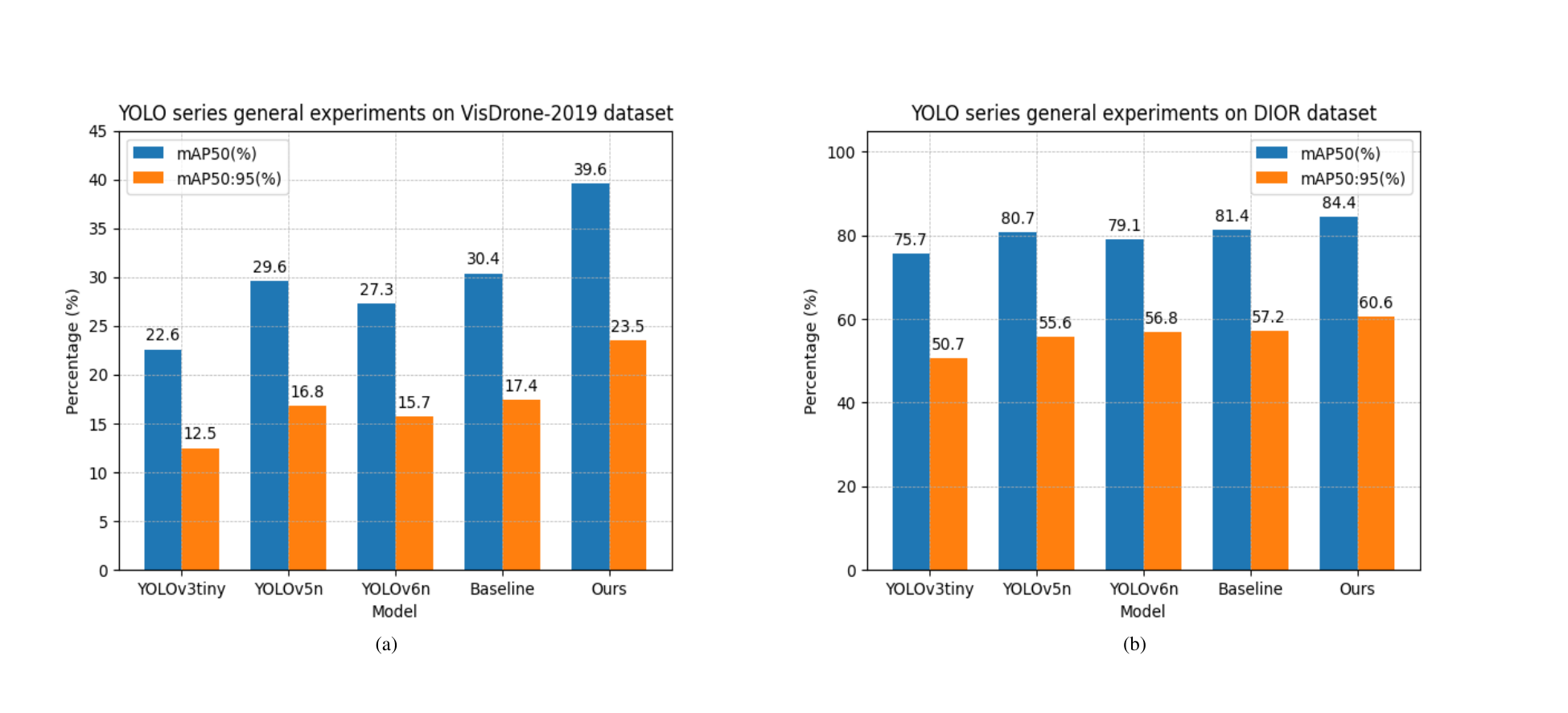}
\vspace{-0.5cm}
\caption{Results of DASSF method in different versions of YOLO model.}
\label{fig:通用性}
\end{figure}

We also apply the finally proposed DASSF method to YOLOv3tiny, YOLOv5n, and YOLOv6n on two datasets. Fig \ref{fig:通用性} shows the experimental results, which shows that the proposed DASSF method is universal.

\subsection{Visualization}  
\vspace{-0.3cm}
\begin{figure}[t]
\centering
\includegraphics[width=12cm]{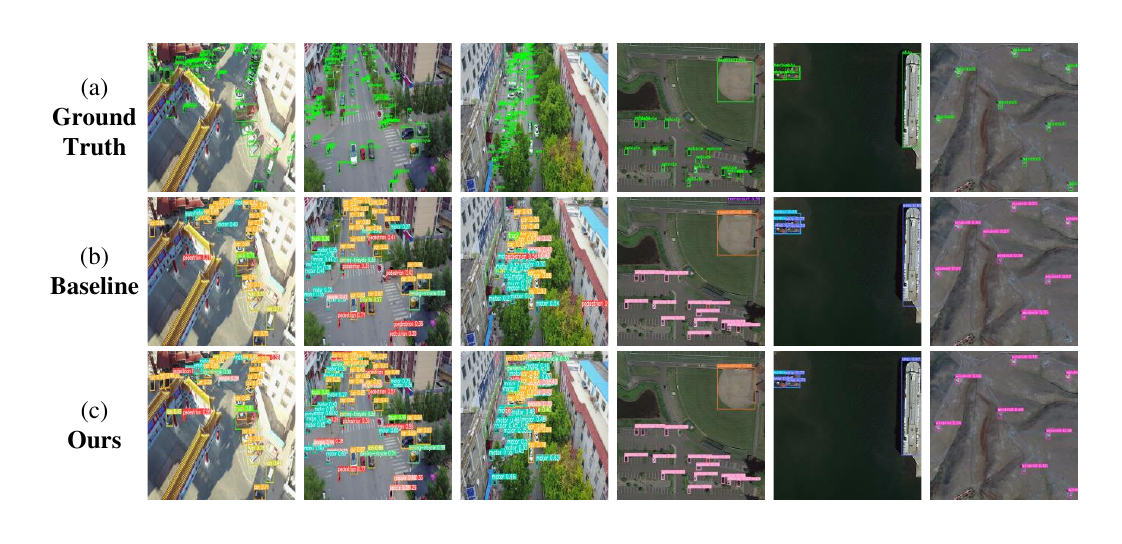}
\caption{Comparison of detection results on VisDrone-2019 and DIOR datasets.}
 
\label{fig:visualization}
\end{figure}

By analyzing the visualization results of the two datasets in fig \ref{fig:visualization}, we find that the DASSF-YOLOv8 model can not only improve the accuracy of densely overlapping objects such as motorcyclists and ports. It can also detect small targets blocked by houses and trees in pictures, as well as targets in blurred images, reducing the missed detection rate. Finally, it also accurately distinguishes between positive and negative samples in the picture to reduce the false detection rate.
 \vspace{-0.3cm}

\section{Conclusion}\label{sec:con} 
\vspace{-0.3cm}
This study proposed an effective aerial image detection method based on dynamic-attention scale- sequence fusion, which improves the problem of low detection accuracy of small targets in aerial images and can be flexibly applied to different YOLO models. We proposed DSSFF module to reduce the amount of calculation. By incorporating additional x-small detection heads, the detection capability of small targets can be improved. Enhanced expression capabilities for various types of targets through the use of DyHead.  Compared with the baseline method or other currently mainstream detection methods, this method improved detection accuracy in all aspects. However, our method still has the drawback of not being fast enough during inference. In the future, we will continue to study how to further lightweight our method so that it can be applied to actual task scenarios.

{\tiny
\bibliography{ref}

\begin{thebibliography}{10}

\bibitem{cai2018cascade}
Zhaowei Cai and Nuno Vasconcelos.
\newblock Cascade r-cnn: Delving into high quality object detection.
\newblock In {\em Proceedings of the IEEE conference on computer vision and pattern recognition}, pages 6154--6162, 2018.

\bibitem{chen2023yolo}
Yuming Chen, Xinbin Yuan, Ruiqi Wu, Jiabao Wang, Qibin Hou, and Ming-Ming Cheng.
\newblock Yolo-ms: rethinking multi-scale representation learning for real-time object detection.
\newblock {\em arXiv preprint arXiv:2308.05480}, 2023.

\bibitem{dai2021dynamic}
Xiyang Dai, Yinpeng Chen, Bin Xiao, Dongdong Chen, Mengchen Liu, Lu~Yuan, and Lei Zhang.
\newblock Dynamic head: Unifying object detection heads with attentions.
\newblock In {\em Proceedings of the IEEE/CVF conference on computer vision and pattern recognition}, pages 7373--7382, 2021.

\bibitem{du2019visdrone}
Dawei Du, Pengfei Zhu, Longyin Wen, Xiao Bian, Haibin Lin, Qinghua Hu, Tao Peng, Jiayu Zheng, Xinyao Wang, Yue Zhang, et~al.
\newblock Visdrone-det2019: The vision meets drone object detection in image challenge results.
\newblock In {\em Proceedings of the IEEE/CVF international conference on computer vision workshops}, pages 0--0, 2019.

\bibitem{jiao2023dsam}
Yuejun Jiao, Song Qiu, Mingsong Chen, Dingding Han, Qingli Li, and Yue Lu.
\newblock Dsam-gn: Graph network based on dynamic similarity adjacency matrices for vehicle re-identification.
\newblock In {\em Pacific Rim International Conference on Artificial Intelligence}, pages 353--364. Springer, 2023.

\bibitem{kang2023asf}
Ming Kang, Chee-Ming Ting, Fung~Fung Ting, and Rapha{\"e}l C-W Phan.
\newblock Asf-yolo: A novel yolo model with attentional scale sequence fusion for cell instance segmentation.
\newblock {\em arXiv preprint arXiv:2312.06458}, 2023.

\bibitem{li2023yolov6}
Chuyi Li, Lulu Li, Yifei Geng, Hongliang Jiang, Meng Cheng, Bo~Zhang, Zaidan Ke, Xiaoming Xu, and Xiangxiang Chu.
\newblock Yolov6 v3. 0: A full-scale reloading.
\newblock {\em arXiv preprint arXiv:2301.05586}, 2023.

\bibitem{li2024lr}
Hanqian Li, Ruinan Zhang, Ye~Pan, Junchi Ren, and Fei Shen.
\newblock Lr-fpn: Enhancing remote sensing object detection with location refined feature pyramid network.
\newblock {\em arXiv preprint arXiv:2404.01614}, 2024.

\bibitem{li2020object}
Ke~Li, Gang Wan, Gong Cheng, Liqiu Meng, and Junwei Han.
\newblock Object detection in optical remote sensing images: A survey and a new benchmark.
\newblock {\em ISPRS journal of photogrammetry and remote sensing}, 159:296--307, 2020.

\bibitem{lin2017focal}
Tsung-Yi Lin, Priya Goyal, Ross Girshick, Kaiming He, and Piotr Doll{\'a}r.
\newblock Focal loss for dense object detection.
\newblock In {\em Proceedings of the IEEE international conference on computer vision}, pages 2980--2988, 2017.

\bibitem{lin2014microsoft}
Tsung-Yi Lin, Michael Maire, Serge Belongie, James Hays, Pietro Perona, Deva Ramanan, Piotr Doll{\'a}r, and C~Lawrence Zitnick.
\newblock Microsoft coco: Common objects in context.
\newblock In {\em Computer Vision--ECCV 2014: 13th European Conference, Zurich, Switzerland, September 6-12, 2014, Proceedings, Part V 13}, pages 740--755. Springer, 2014.

\bibitem{liu2016ssd}
Wei Liu, Dragomir Anguelov, Dumitru Erhan, Christian Szegedy, Scott Reed, Cheng-Yang Fu, and Alexander~C Berg.
\newblock Ssd: Single shot multibox detector.
\newblock In {\em Computer Vision--ECCV 2016: 14th European Conference, Amsterdam, The Netherlands, October 11--14, 2016, Proceedings, Part I 14}, pages 21--37. Springer, 2016.

\bibitem{liu2023learning}
Wenze Liu, Hao Lu, Hongtao Fu, and Zhiguo Cao.
\newblock Learning to upsample by learning to sample.
\newblock In {\em Proceedings of the IEEE/CVF International Conference on Computer Vision}, pages 6027--6037, 2023.

\bibitem{lyu2022rtmdet}
Chengqi Lyu, Wenwei Zhang, Haian Huang, Yue Zhou, Yudong Wang, Yanyi Liu, Shilong Zhang, and Kai Chen.
\newblock Rtmdet: An empirical study of designing real-time object detectors.
\newblock {\em arXiv preprint arXiv:2212.07784}, 2022.

\bibitem{ni2018multiple}
Haomiao Ni, Hong Liu, Zichao Guo, Xiangdong Wang, Taijiao Jiang, Kuansong Wang, and Yueliang Qian.
\newblock Multiple visual fields cascaded convolutional neural network for breast cancer detection.
\newblock In {\em PRICAI 2018: Trends in Artificial Intelligence: 15th Pacific Rim International Conference on Artificial Intelligence, Nanjing, China, August 28--31, 2018, Proceedings, Part I 15}, pages 531--544. Springer, 2018.

\bibitem{qiao2022novel}
Chenchen Qiao, Fei Shen, Xuejun Wang, Ruixin Wang, Fang Cao, Sixian Zhao, and Chang Li.
\newblock A novel multi-frequency coordinated module for sar ship detection.
\newblock In {\em 2022 IEEE 34th International Conference on Tools with Artificial Intelligence (ICTAI)}, pages 804--811. IEEE, 2022.

\bibitem{shen2023pbsl}
Fei Shen, Xiangbo Shu, Xiaoyu Du, and Jinhui Tang.
\newblock Pedestrian-specific bipartite-aware similarity learning for text-based person retrieval.
\newblock In {\em Proceedings of the 31th ACM International Conference on Multimedia}, 2023.

\bibitem{shen2023git}
Fei Shen, Yi~Xie, Jianqing Zhu, Xiaobin Zhu, and Huanqiang Zeng.
\newblock Git: Graph interactive transformer for vehicle re-identification.
\newblock {\em IEEE Transactions on Image Processing}, 2023.

\bibitem{wang2024gold}
Chengcheng Wang, Wei He, Ying Nie, Jianyuan Guo, Chuanjian Liu, Yunhe Wang, and Kai Han.
\newblock Gold-yolo: Efficient object detector via gather-and-distribute mechanism.
\newblock {\em Advances in Neural Information Processing Systems}, 36, 2024.

\bibitem{wang2019carafe}
Jiaqi Wang, Kai Chen, Rui Xu, Ziwei Liu, Chen~Change Loy, and Dahua Lin.
\newblock Carafe: Content-aware reassembly of features.
\newblock In {\em Proceedings of the IEEE/CVF international conference on computer vision}, pages 3007--3016, 2019.

\bibitem{wang2023improved}
Xin Wang, Ning He, Chen Hong, Qi~Wang, and Ming Chen.
\newblock Improved yolox-x based uav aerial photography object detection algorithm.
\newblock {\em Image and Vision Computing}, 135:104697, 2023.

\bibitem{weng2023cross}
Weijie Weng, Weiming Ling, Feng Lin, Junchi Ren, and Fei Shen.
\newblock A novel cross frequency-domain interaction learning for aerial oriented object detection.
\newblock In {\em Chinese Conference on Pattern Recognition and Computer Vision (PRCV)}. Springer, 2023.

\bibitem{weng2024enhancing}
Weijie Weng, Mengwan Wei, Junchi Ren, and Fei Shen.
\newblock Enhancing aerial object detection with selective frequency interaction network.
\newblock {\em IEEE Transactions on Artificial Intelligence}, 1(01):1--12, 2024.

\bibitem{xu2022damo}
Xianzhe Xu, Yiqi Jiang, Weihua Chen, Yilun Huang, Yuan Zhang, and Xiuyu Sun.
\newblock Damo-yolo: A report on real-time object detection design.
\newblock {\em arXiv preprint arXiv:2211.15444}, 2022.

\bibitem{zhang2020bridging}
Shifeng Zhang, Cheng Chi, Yongqiang Yao, Zhen Lei, and Stan~Z Li.
\newblock Bridging the gap between anchor-based and anchor-free detection via adaptive training sample selection.
\newblock In {\em Proceedings of the IEEE/CVF conference on computer vision and pattern recognition}, pages 9759--9768, 2020.

\bibitem{zhao2018surface}
Zhixuan Zhao, Bo~Li, Rong Dong, and Peng Zhao.
\newblock A surface defect detection method based on positive samples.
\newblock In {\em PRICAI 2018: Trends in Artificial Intelligence: 15th Pacific Rim International Conference on Artificial Intelligence, Nanjing, China, August 28--31, 2018, Proceedings, Part II 15}, pages 473--481. Springer, 2018.

\end{thebibliography}
}
\end{document}